\pdfoutput=1
\documentclass[11pt]{article}

\usepackage[]{acl}

\usepackage{tikz}

\usepackage{times}
\usepackage{latexsym}

\usepackage[T1]{fontenc}
\usepackage[utf8]{inputenc}

\usepackage{microtype}

\usepackage{xcolor, soul}
\usepackage{graphicx}
\usepackage{subfigure}
\usepackage{booktabs, multirow} % for borders and merged ranges
\usepackage{soul}% for underlines
\usepackage{amsmath}
\usepackage{algpseudocode}
\usepackage{makecell}
\usepackage{balance}
\usepackage{listings}
\usepackage{amsmath}
\usepackage{amssymb}
\usepackage{tcolorbox}

\definecolor{codegreen}{rgb}{0,0.6,0}
\definecolor{codegray}{rgb}{0.5,0.5,0.5}
\definecolor{codepurple}{rgb}{0.58,0,0.82}
\definecolor{backcolour}{rgb}{0.95,0.95,0.92}
\lstdefinestyle{mystyle}{
    backgroundcolor=\color{backcolour},   
    commentstyle=\color{codegreen},
    keywordstyle=\color{magenta},
    numberstyle=\tiny\color{codegray},
    stringstyle=\color{codepurple},
    basicstyle=\ttfamily\footnotesize,
    breakatwhitespace=false,         
    breaklines=true,                 
    captionpos=b,                    
    keepspaces=true,                 
    numbers=left,                    
    numbersep=5pt,                  
    showspaces=false,                
    showstringspaces=false,
    showtabs=false,                  
    tabsize=2
}
\newcommand{\ours}{\textsc{KGLens}}

\NewDocumentCommand\emojiLensTitle{}{
    \raisebox{-0.2\height}{\includegraphics[scale=4]{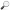}}
}
\title{KGLens\emojiLensTitle: Towards Efficient and Effective Knowledge Probing of \\Large Language Models with Knowledge Graphs}
\author{Shangshang Zheng\thanks{~~Equal contribution.}, He Bai\footnotemark[1], Yizhe Zhang, Yi Su,\\
{\bf Xiaochuan Niu, Navdeep Jaitly} \\
Apple \\
{\tt \{dan.z, hbai22, yizhe\_zhang, yi\_su, xniu, njaitly\}@apple.com}
}

\begin{document}
\maketitle

\begin{abstract}
Large Language Models~(LLMs) might hallucinate facts, while curated Knowledge Graph~(KGs) are typically factually reliable especially with domain-specific knowledge. 
Measuring the alignment between KGs and LLMs can effectively probe the factualness and identify the knowledge blind spots of LLMs. 
However, verifying the LLMs over extensive KGs can be expensive.
In this paper, we present \ours, a Thompson-sampling-inspired framework aimed at effectively and efficiently measuring the alignment between KGs and LLMs.
%, and pinpointing the LLMs' knowledge deficiencies relative to KGs. 
\ours~features a graph-guided question generator for converting KGs into natural language, along with a carefully designed importance sampling strategy based on parameterized KG structure to expedite KG traversal.
Our simulation experiment compares the brute force method with \ours~under six different sampling methods, demonstrating that our approach achieves superior probing efficiency. 
Leveraging \ours, we conducted in-depth analyses of the factual accuracy of ten LLMs across three large domain-specific KGs from Wikidata, composing over 19K edges, 700 relations, and 21K entities. 
Human evaluation results indicate that \ours~can assess LLMs with a level of accuracy nearly equivalent to that of human annotators, achieving 95.7\% of the accuracy rate.
\end{abstract}

\section{Introduction}
% The factualness of Large Language Models~(LLMs) is of paramount importance for their reliability and utility in various applications. 
The factualness of Large Language Models~(LLMs) is crucial for their reliability and utility in various applications.
Nonetheless, studies have shown that LLMs can produce information that is nonfactual, hallucinated, or outdated~\cite{perez2022red,ji2023survey,lee2022factuality,wang2021can}.

To evaluate the factualness of LLMs, fact-checking~\cite{thorne2018fever,augenstein2023factuality} and fact-QA~\cite{petroni2020kilt, press2022measuring, dhingra-etal-2022-time} approaches are commonly used, but several challenges persist.
For fact-checking, distinguishing faithful and unfaithful statements is different from evaluating factualness of the generation.
For fact-QA, scaling up the evaluation data is challenging due to the expensive nature of the annotation process.
For both approaches, it is hard to exclude their data from the web-crawled pretraining corpus~\cite{deng2023benchmark} to ensure the fairness of the evaluation.
Last but not least, LLMs may respond differently to the same fact when the quert is phrased in different forms, a challenge that existing fact-checking and fact-QA datasets are not equipped to handle.

In contrast, probing LLMs knowledge by transforming knowledge graph~(KG) data into natural language questions addresses these limitations. 
First, KGs are inherently scalable, factually reliable, and tailored to specific domains. 
This scalability allows for extensive and efficient evaluation of LLMs across various domains. 
% The factual reliability of KGs ensures that the evaluation data is accurate and dependable, leading to more valid assessments of LLMs' factual correctness.
Second, question generation~(QG) from KGs can be automated, enabling the rapid and large-scale creation of evaluation datasets.
Moreover, generating various questions from the same set of facts allows us to assess the robustness of LLMs to different phrasings.

However, there are several challenges for QG-based KG evaluation. 
The first is the efficiency issue.
Existing knowledge probing methods~\cite{dong2023statistical, wang2023assessing}, mainly using the brute-force method which exhaustively evaluates all KG edges, are computationally expensive, time-consuming, and non-scalable.
The second challenge comes to transformation of ambiguous KG triplets into natural language queries.
Given a subject and predicate, the correct object may not be unique, thus the answer to the question is ambiguous.
Furthermore, some subjects may lack specificity, resulting in confusing questions. 
Examples illustrating these issues are shown in Table.~\ref{tab:triple_examples}.
In this case, both text-cloze based methods~\cite{petroni2019language,jiang2020can,dong2023statistical} and prompt-based QG methods~\cite{wang2023assessing} fail to evaluate the factualness properly. 

\begin{table}[!t]
    \centering
    \scriptsize
    \begin{tabular}{l} 
    \toprule
        \textbf{Multi-objects}\\
    \hline
    Barack Obama, educated at, Columbia University \\
    Barack Obama, educated at, Harvard Law School \\
    \hline
    \textbf{Ambiguous subject} \\
    Waterloo (also known as Waterloo, Ontario) \\
    Waterloo (also known as Waterloo, Iowa) \\
    \bottomrule
    \end{tabular}
    \caption{Examples of ambiguous KG triplets.}
    \label{tab:triple_examples}
\end{table}

% \citet{petroni2019language} proposed to transform KG triplet into text-cloze task but the formulated sentences are ambiguous and unnatural.
% \citet{jiang2020can} alleviate this issue by mining the relation words from the web for each subject-object pair, which is impractical for large graph.
% Another challenge is the efficiency of the evaluation.
% KGs are typically large. 
% And evaluating the robustness of an LLM's knowledge may necessitate multiple evaluation rounds using the same KG, as an LLM may respond differently to the same query.

% Moreover, these approaches neglect additional information within the KG, including the graph structure, node attributes, predicate attributes, and more, all of which play a crucial role in achieving more accurate knowledge retrieval and reducing the ambiguity of KG edges.

\begin{figure*}[!ht]
\centering
\includegraphics[width=0.55\textwidth]{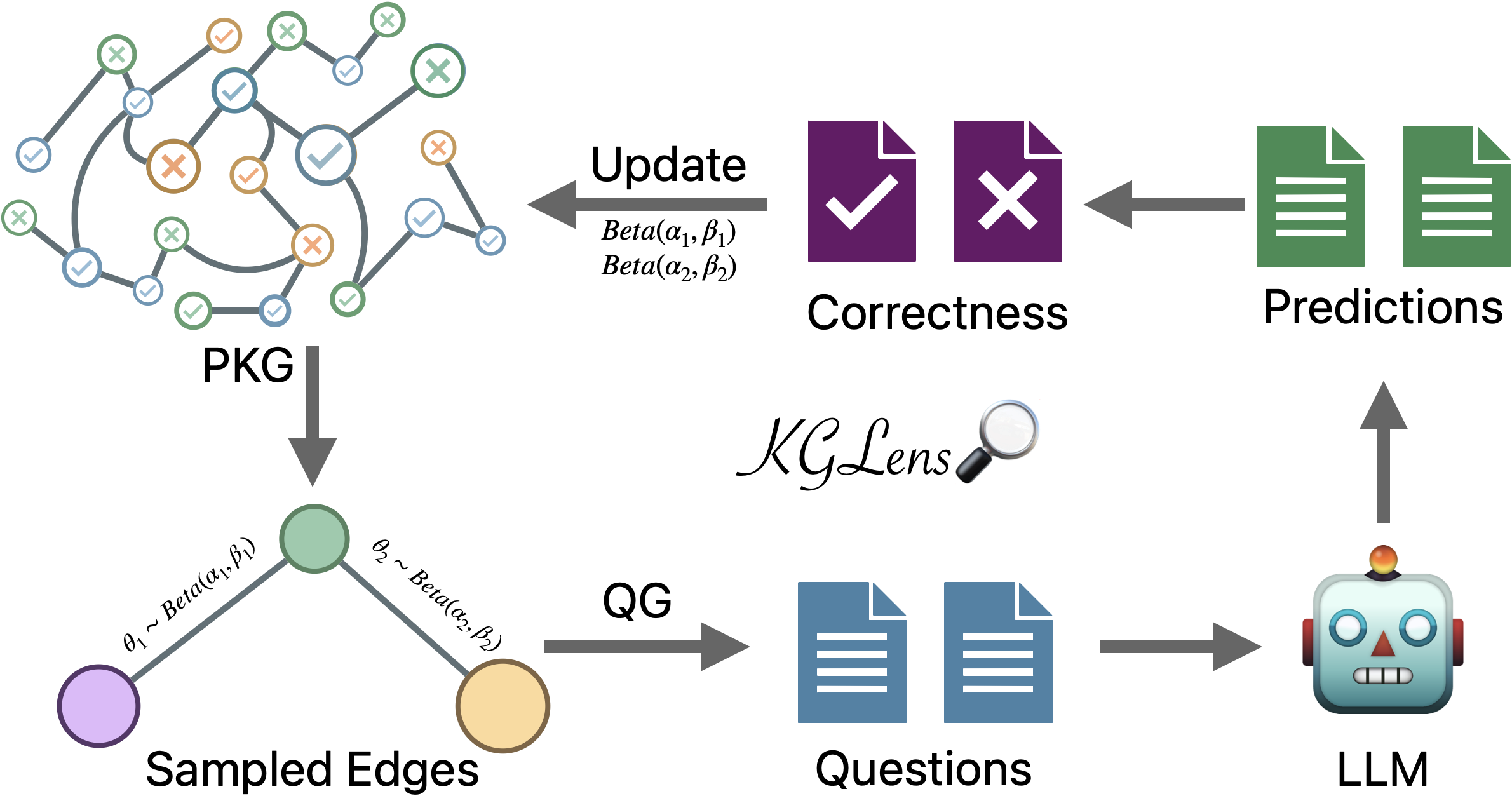}
\caption{\ours~Framework. \ours~starts from the PKG initialization, where each edge is augmented with a beta distribution. Then a batch of edges is sampled based on the edge probability $\theta$. After that, questions are generated from these edges and an LLM will be examined with question answering task. Then we update the beta distribution of PKG edges based on the QA results. We iterate this process until the running metrics are converged.}
\label{fig:arch}
\end{figure*}

%Richard: will revise the context below later

% A comments for Intro section from Daniel:
% Currently in the intro we first mentioned challenge of transforming KG into natural language and then talked about the evaluation efficiency, should switch the order?

In this study, we present a novel knowledge probing framework named \ours~(Fig.~\ref{fig:arch}), to measure the knowledge alignment between KGs and LLMs, and to identify LLM's \textit{knowledge blind spots}.
To efficiently probe the LLM, \ours~features a Thompson sampling inspired method to rapidly identify knowledge blind spots. 
More specifically, we first introduce a parameterized knowledge graph~(PKG), where each KG edge is augmented with a beta distribution, serving as an indicator of the LLM's deficiency on that specific edge. 
We apply Thompson sampling to select the edges based on edge deficiency and then evaluate LLMs with the sampled edges, update the PKG with the evaluation results, and iterate this process until the running metrics converge.
Our simulation experiments show that \textit{our sampling method with PKG is more efficient than random sampling and brute-force methods. }
% In this way, when an LLM is unable to provide a satisfactory response to a question, the KG structure enables us to pinpoint the relevant source edge and entities. 
% This information can then be used to update the PKG, and the process can be iteratively applied until adequate coverage is achieved.

To accurately probe the LLM, \ours~features a graph-guided question generator for converting KGs into natural language with GPT-4~\cite{OpenAI2023GPT4TR}.
We design two types of questions~(fact-checking and fact-QA) to reduce the ambiguity of the expected answer, where the question type is controlled by the graph structure.
We also include the entity aliases during the question generation to provide additional context and reduce the subject ambiguity.
\textit{Human evaluation results show that 97.7\% of our generated questions are sensible to human annotators.}

To assess the whole framework of~\ours, we probed 10 popular LLMs with three domain-specific KGs collected from Wikidata, encompassing over 700 relations and 21K entities.
We introduce three evaluation metrics to measure the knwoledge alignment between an LLM and a KG: zero-sense rate~(percentage of facts that an LLM never answered correctly in all rounds), all-sense rate~(percentage of facts that an LLM always answered correctly in all rounds), and win rate~(percentage of facts that an LLM answered correctly more than half of the rounds).
% Our results show that \ours~is capable not only of accessing the factualness of LLMs but also of pinpoints LLMs' knowledge deficiencies relative to KGs in different levels~(e.g., temporal and topics).
Human evaluation shows that \ours~can assess LLMs with a level of accuracy nearly equivalent to that of human annotators with 95.7\% accuracy rate. 
Our key contributions are as follows:
\begin{enumerate}
    \item We introduce a novel, efficient knowledge probing framework to efficiently identify LLMs' knowledge blind spots across diverse topics and relationships.
    \item Our QG strategy leverages KG structure to reduce ambiguity thus providing accurate and effective LLM evaluation, surpassing existing methods that focus solely on atomic triplets.
    \item Human evaluation verifies our framework's efficacy with a human-level accuracy (95.7\%) in assessing LLMs' knowledge, nearly matching human annotator performance.
    \item We propose three metrics to quantify knowledge alignment between LLMs and KGs, successfully identifying knowledge gaps and reliable facts.
    \item Our contributions advance the development of more reliable and factual AI systems, promoting trustworthy user experiences and efficient model improvement. We will open-source our framework and datasets.
\end{enumerate}

% In our experiments, the iteratively updated parameterized KG represents the LLM's knowledge proficiency over a selected KG.
% By preserving the KG structure, we are empowered to conduct highly adaptable and customizable analysis, including factors such as entity types/groups, predicate types, and temporal aspects.
% Our contributions are as follows:
% \begin{itemize}
%     \item We propose \ours, an efficient method for visualizing and assessing the factual knowledge contained in LLMs. 
%     \ours~yields highly adaptable and customizable views of the LLM's knowledge by leveraging the KG structure.
%     \item The proposed parameterized KG allows us to efficiently evaluate the knowledge reliability of LLMs.
%     \item The proposed graph-guided QG strategy enables us to evaluate LLMs in a way that is more similar to human interaction.
%     \item We have developed three domain-specific KGs from Wikidata, encompassing over 700 relations and 21K entities. 
%     These KGs will be released for future research.
% \end{itemize}

\section{Method}

In probing the knowledge of LLMs, we face a unique challenge: while we have access to a vast corpus of KG edges, we lack insight into how well the LLM understands each of them. 
This scenario bears a striking resemblance to the multi-armed bandit problem in reinforcement learning, where we must balance exploration and exploitation to maximize our understanding of the LLM's knowledge gaps.

Our approach, which we call~\ours, aims to efficiently identify the edges where the LLM's knowledge is weakest or most limited. This allows us to focus our efforts on the areas where the LLM needs the most improvement. 
The framework of~\ours~is illustrated in Fig.~\ref{fig:arch}, with its workflow summarized in the caption.
Key aspects of our method include Thompson Sampling Inspired Parameterized KG, Graph-Guided Question Generator, and Answer Verification.

In the following subsections, we will delve into each of these components in detail, explaining how they work together to create an effective knowledge probing system for LLMs.

\subsection{Parameterized Knowledge Graph}
\label{sec:pkg}
A knowledge graph $\mathcal{G}$, is a set of triplets $\{(s_j, p_j, o_j)\}_{j=1\cdots i}$ where each tuple describes a relationship (predicate) $p_j$ between a subject $s_j$ and an object $o_j$. 

% Efficient KG traversal is necessary as KG can be enormous, and revisiting the edges that the LLM respond incorrectly is important to get a clear picture of an LLM's knowledge. 
Intuitively, if an LLM failed in answering a question, there is a higher chance that the LLM also lacks knowledge of the related topics. To reflect this intuition, we propose a parameterized KG (PKG), by augmenting each edge $(s_j, p_j, o_j)$ with an additional error probability $\theta_j$ reflecting the probability that an LLM may fail on this edge. 
We use a Beta distribution to model $\theta$:
\begin{equation}
    \theta_j \sim Beta(\alpha_j, \beta_j),
\end{equation}
where $\alpha$ and $\beta$ can be interpreted as the number of times the targeted LLM failed or succeeded in answering the question. The prior of each $\theta_j$ is set to $Beta(1,1)$.

The estimation of the posterior $\{\alpha_j, \beta_j\}_{\forall j}$ is done in an iterative manner based on the outcome from the LLM. 
Each iteration consists two stages: edge sampling and parameter updating. 

\subsubsection*{Edge sampling} The edge sampling process favors the edges with larger $\theta$ values.
In each iteration, we sample top-n edges ranked by $\theta$ sampled from the Beta distributions of PKG, and then send these edges to LLM for examination and verification. 
The signal regarding the correctness of LLM's response is collected for each edge.

\subsubsection*{Parameter estimation and updating}
For each edge, the $\alpha$ and $\beta$ is updated based on the new observation of whether the response from LLM is correct, following the standard Beta distribution posterior updates.
In addition, we also propagate the signal to the neighboring edges to account for the high correlation in error probability among the connected edges.
%In order to account for the high correlation in error probability among the connected edges, we have additionally propagate the signal to the neighboring edges. Specifically, the signal gathered from $p_j$ is propagated to both the incoming and outgoing edges that are connected to node $s_j$ and $o_j$. 
To optimize the computational process, signal propagation is restricted to one degree. Specifically, 
\begin{align}
\alpha_j &= \alpha_j + \mathbb{I}(\text{response is incorrect}) + M_j, \\
\beta_j &= \beta_j + \mathbb{I}(\text{response is correct}) + N_j, 
\end{align}
where $M_j = \lvert \text{incorrect neighborhood edges}\rvert$ and $N_j = \lvert \text{correct neighborhood edges}\rvert$.
An updated PKG~(Fig.~\ref{fig:pkg}) is then obtained by repeating the edge sampling and parameter updating process iteratively until the running metrics~(\ref{sec:metrics}) converge.

\subsection{Graph-guided Question Generation} \label{subsec:question-generation}
We use GPT-4 to transform the sampled edge $K_i$ into the natural questions with few-shot in-context learning.
The prompts and demonstrations are shown in Appendix~\ref{subsec:prompt}.
We design two types of questions for \ours: Yes/No Questions~(judgement) and Wh-Questions~(generative), where the question type is controlled by the graph structure~(out degree).
In addition, to reduce the ambiguity of entities, we provide the entity alias for question generation.

\subsubsection{Yes/No Questions} \label{subsec:yes-no}
Each KG edge can be transformed into a question by asking if the subject's relation is the object.
But in this way, the answer would always be \textit{Yes} for all the edges.
To formulate hard negative examples, we build a ground truth answer set $\mathbf{T_j}$ for each $(s_j, p_j)$, and the candidate answer set $\mathbf{C_j}$ for each $p_j$. 
Both $\mathbf{T_j}$ and $\mathbf{C_j}$ are derived from the full Wikidata knowledge graph to ensure the completeness.
Then, for a tuple $\{(s_j, p_j, o_j)\}$, we use $o_j$ to constitute the \textit{Yes} question, and sample a random $o_x$ from $\mathbf{C_j} - \mathbf{T_j}$ to formulate the \textit{No} question.
Considering our QG process is on-the-fly during the evaluation, \ours~can formulate different QA pairs for the same fact.
The sampling rate between yes and no question is evenly split, with a 50-50 distribution.

\subsubsection{Wh-Questions} \label{subsec:generative-question}
Another type of question queries the LLMs to generate the object(s) given the subject and the predicate, where the questions usually begin with when/where/who/what.
While this question type presents greater difficulty, it is not universally applicable. Wh-questions may yield hundreds of correct objects, rendering exhaustive enumeration impractical and uninformative.

In \ours, we opt to generate Wh-Questions only when the out degree of an entity is less than 10.
Otherwise, the Yes/No Questions prompt is adopted.

\begin{figure}[!t]
  \centering
  \includegraphics[width=0.3\textwidth]{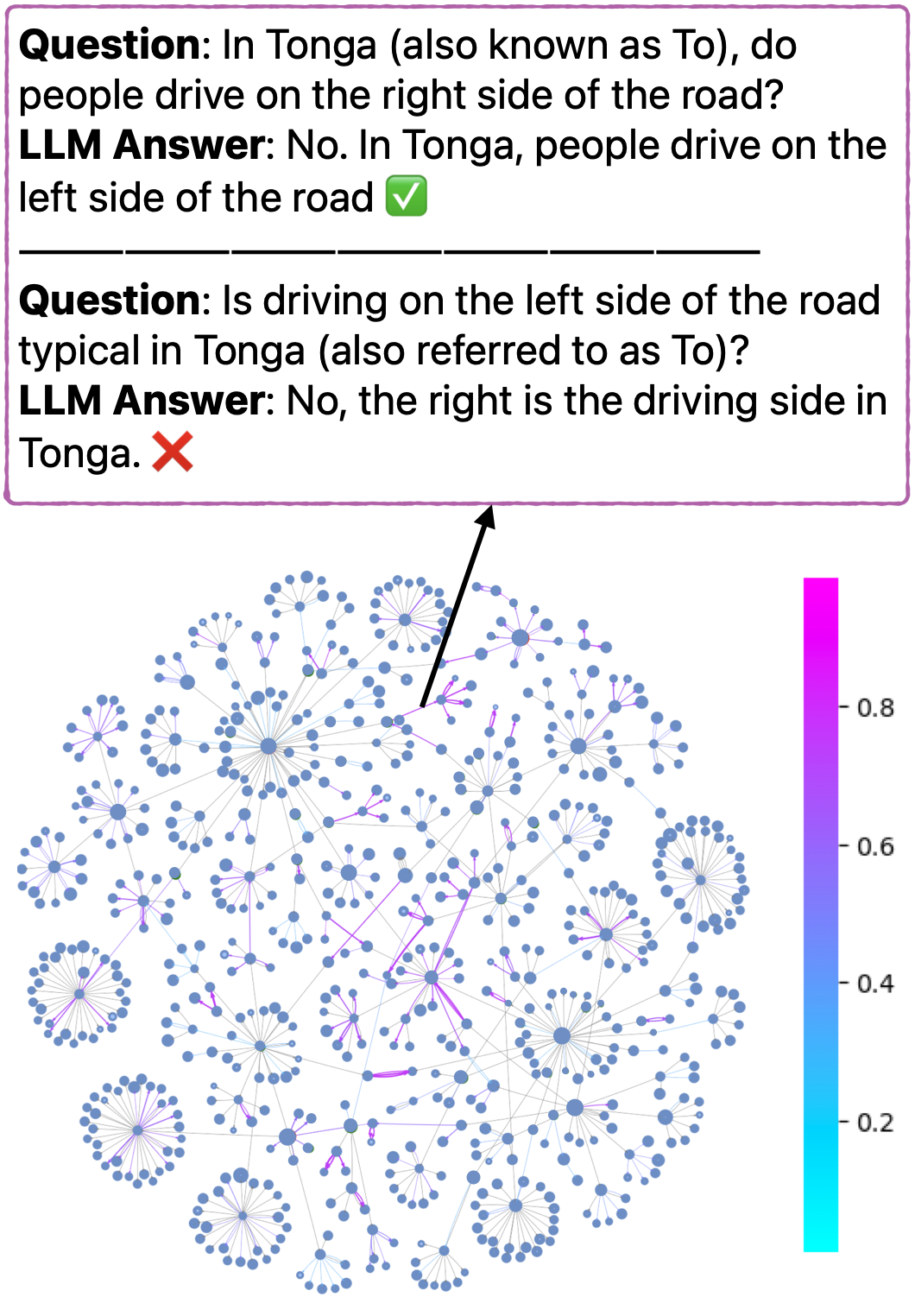}
  \caption{Parameterized KG. The edge color is an estimation in LLM's deficiency to the associated fact.}
  \label{fig:pkg}
  \end{figure}

\subsection{QAV: Question Answering Verification}
Our QA testing involved two difficulty levels: EASY, comprising solely Yes/No questions; and HARD, with each question type generated at a 50\% chance. Few-shot in-context learning was employed to test the LLMs.

To verify the response, we guide the LLMs to generate either ``Yes'' or ``No'' at the beginning of the response for Yes/No Questions and subsequently generate accompanying explanations. 
This approach facilitates a straightforward verification process by examining the correspondence of the initial word.
For Wh-Questions, we instruct the LLM to list all the correct answers.
In this case, the assessment of the answer cannot be done by string matching. 
Therefore, we employ a GPT-4 model to check the correctness of a response given the question, the ground truth objects and their aliases. 
The prompts are listed in Appendix~\ref{subsec:prompt}.

\begin{figure*}[!ht]
  \centering
  \begin{subfigure}[]
    {\includegraphics[width=0.45\textwidth]{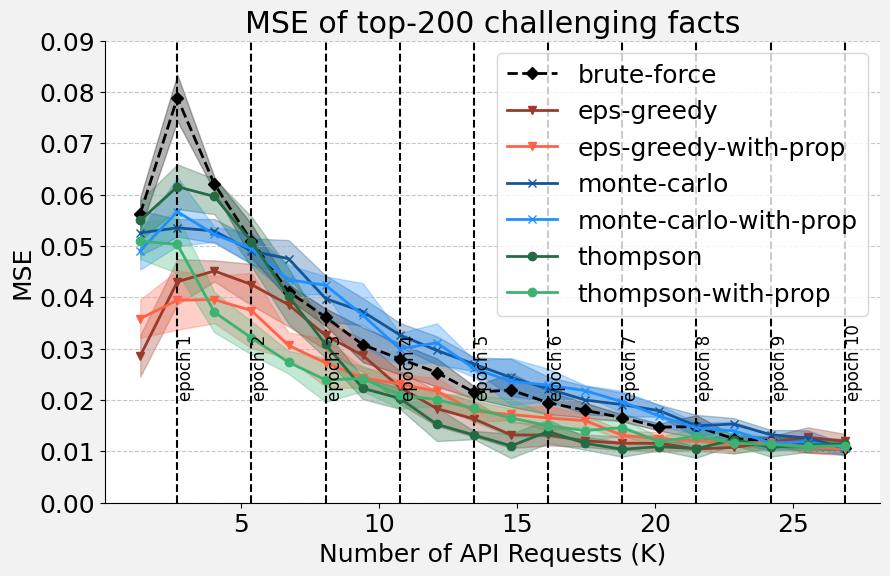}\label{fig:top200_mse_over_compute}}
  \end{subfigure}
  \begin{subfigure}[]
  {\includegraphics[width=0.45\textwidth]{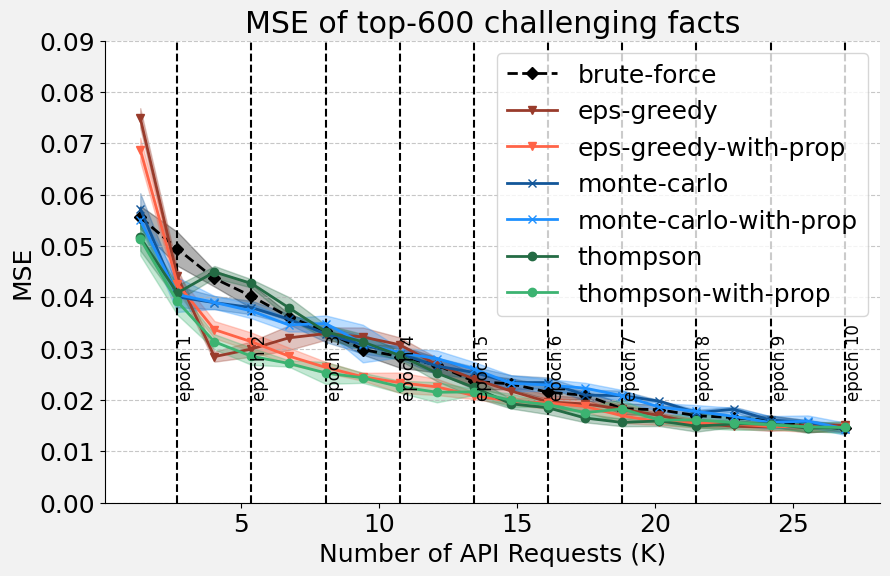} \label{fig:top600_mse_over_compute}}
  \end{subfigure}
  % \begin{subfigure}[]
  % {\includegraphics[width=0.35\textwidth]{figs/mse_of_top-1200_challenging_facts.png} \label{fig:top1200_mse_over_compute}}
  % \end{subfigure}
  % \begin{subfigure}[]
  % {\includegraphics[width=0.35\textwidth]{figs/mse_of_top-1800_challenging_facts.png} \label{fig:top1800_mse_over_compute}}
  % \end{subfigure}
  \caption{We measure the MSE distance between the ground truth $\theta$ and the estimated $\theta'$ across different sampling method. The vertical line epoch-N shows the number of API requests required for the Brute Force method to complete N full iterations over each edges. 
  %Fig.~\ref{fig:top200_mse_over_compute} and Fig.~\ref{fig:top600_mse_over_compute} show how different methods perform in terms of accurately evaluating the top-200 and top-600 challenging edges. 
  The graph consists of 2789 edges in total.}%Thus, these charts demonstrate how each method uncovers blind spots in a 2689-edge-sized knowledge graph.}
  \label{fig:rpms}
  \end{figure*}

\subsection{Evaluation Efficiency Study}
In this section, we conduct a simulation experiment to evaluate the efficiency of different methods in identifying the most challenging edges in a domain specific KG~(NBA KG in Table~\ref{tab:graph-stats}), including: Thompson sampling, Epsilon Greedy, Monte Carlo, and the straightforward iteration method (Brute Force), which involves iterating over all edges multiple times. 
We first estimate the ground truth $\theta$ by iterating over all edges 20 times using the Brute-Force method.
Then, top-k difficult edges can be identified by ranking the estimated $\theta$.
Since we are interested in the most challenging edges, Mean Square Error (MSE) for these top-k edges will be computed between the estimated $\theta$ and the ground truth $\theta$ for each method. 

In Fig.~\ref{fig:rpms}, we plot different methods' MSE score of top-200 and top-600 challenging edges over the number of API requests.
The vertical epoch lines represent the cumulative number of API requests necessary to complete a full iterations over the PKG edges using the Brute-Force method.
From this figure, we observe Thompson sampling method demonstrates superiority in pinpointing the most challenging edges. 
Notably, both Thompson sampling and Epsilon-Greedy exhibit advantages over Monte Carlo and Brute-Force methods by striking an effective balance between exploration and exploitation. 
For exmaple, from Fig.~\ref{fig:top600_mse_over_compute}, we can see that Thompson sampling with propogation achieves the accuracy of a 4.5-epoch Brute-Force method while utilizing only 65\% of the compute resources. 
This finding underscores the potential of our approach to swiftly identify critical edges and optimize resource allocation in large-scale KG applications.
In addition, we find the signal propagation mechanism further enhances the capabilities, enabling them to achieve better accuracy rapidly.

\section{Experiments}
\begin{table}[!t]
    \centering
    \scriptsize
    \begin{tabular}{ccccc} 
    \toprule
         KG&  Active Edges& Dead Edges & Nodes&Predicates\\ 
    \hline
         Country&  7844& 9441 & 12760&338\\ 
         NBA&  2689& 1158 & 805&57\\ 
         Movies&  8704& 3053 & 7965&340\\ 
    \bottomrule
    \end{tabular}
    \caption{Statistics of the testing knowledge graphs.}
    \label{tab:graph-stats}
\end{table}

\begin{table*}[!htp]\centering \scriptsize
   \begin{tabular}{lrrrrrrrr}\toprule
    \multirow{2}{*}{LLMs} &\multicolumn{2}{c}{Country} &\multicolumn{2}{c}{NBA} &\multicolumn{2}{c}{Movie} &\multirow{2}{*}{Average} \\\cmidrule{2-7}
   &EASY &HARD &EASY &HARD &EASY &HARD \\\midrule
    Babbage-002 &57.46 &34.39 &58.32 &27.65 &57.48 &31.00 &44.38 \\
    Davinci-002 &58.85 &38.36 &58.21 &30.57 &55.66 &34.72 &46.06 \\
    Vicuna-33b-v1.3 &66.51 &55.87 &36.60 &41.66 &50.56 &46.22 &49.57 \\
    Xwin-LM-13B-V0.2 &54.77 &49.13 &53.52 &50.51 &53.59 &47.84 &51.56 \\
    Yi-34B-Chat &66.72 &56.16 &65.66 &62.06 &59.86 &55.78 &61.04 \\
    GPT-3.5-turbo &74.43 &63.42 &57.98 &56.95 &62.80 &57.70 &62.21 \\
    GPT-4-1106-preview &82.27 &72.42 &79.09 &70.57 &83.15 &66.95 &75.74 \\
    GPT-4-turbo &83.69 &73.19 &82.38 &76.11 &85.62 &73.58 &79.09 \\
    GPT-4o &\textbf{86.25} &73.71 &78.45 &78.46 &\textbf{86.36} &\textbf{74.34} &79.59 \\
    GPT-4 &84.79 &\textbf{74.06} &\textbf{84.23} &\textbf{78.93} &85.14 &70.80 &\textbf{79.66} \\
   \bottomrule
   \end{tabular}
   \caption{Win rate results for different LLMs evaluated under EASY and HARD modes.}\label{tab:win-rate}
\end{table*}

% \begin{table*}[!ht]
%    %\resizebox{1/2\textwidth}{!}{% use resizebox with textwidth
%        \centering
%        \begin{tabular}{cccccc}
%          \hline
%             &  babbage-002&  davinci-002&  gpt-3.5-turbo& gpt-4 &gpt-4-1106-preview\\
%          \hline
%             Country&  57.46&  58.85&  71.43& 84.79 &82.27\\
%             NBA&  58.32&  58.21&  57.98& 84.23 &79.09\\
%             Movie&  57.48&  55.66&  62.80& 85.14 &83.15\\
%          \hline
%        \end{tabular}
%        \caption{EASY win rate. The random guessing baseline is 50\%.}
%        \label{tab:easy-win-rate}
% \end{table*}
% \begin{table*}[!ht]
% \centering
%        \begin{tabular}{cccccc}
%          \hline
%             &  babbage-002&  davinci-002&  gpt-3.5-turbo& gpt-4 &gpt-4-1106-preview\\
%          \hline
%             Country&  24.51&  24.44&  14.98& 7.42 &7.59\\
%             NBA&  15.34&  17.69&  17.17& 6.07 &8.19\\
%             Movie&  26.17&  28.54&  22.7& 8.35 &9.21\\
%          \hline
%        \end{tabular}
%       % }
%        \caption{EASY zero-sense rate}
%        \label{tab:easy-zero-sense}
% \end{table*}

In this paper, we develop three domain-specific KGs using Wikidata to evaluate the knowledge accuracy and reliability of two widely used LLM APIs~(GPT-3.5-turbo and GPT-4), two legacy LLMs~(Babbage-002 and Davinci-002), together with three variations of GPT-4~(GPT-4-1106-preview, GPT-4-turbo, GPT-4o). We also evaluated three open source LLMs including Vicuna-33b-v1.3~\cite{vicuna2023}, Xwin-LM-13B-V0.2~\cite{xwin-lm}, and Yi-34B-Chat \footnote{\url{https://www.01.ai}}. Web browsing ability is unavailable to OpenAI API calls at the time when experiments are conducted.

\subsection{Dataset}
We prepare three test datasets~(KG) with Wikidata Query Web Service \footnote{\url{https://query.wikidata.org}} in three topics: country, NBA, and movie.
The country KG includes knowledge about 16 countries. 
The NBA KG contains the knowledge related to 30 NBA teams. 
And the movies are sampled from films after 2015.

The statistics of our KGs are shown in Tab.~\ref{tab:graph-stats}.
The term \textbf{``dead edges''} refers to edges that are less intriguing to inquire about but are still crucial for displaying entity relations. 
For example, certain predicates such as ``member of'' and ``domestic relation'', exemplify links between entities, but they are less captivating to inquire about and are too prevalent.
Conversely, significant and meaningful edges are referred to as \textbf{``active edges''}, and we use them to generate questions. 
% Active edges represent the essential and noteworthy connections in the knowledge graph, from which we extract information to formulate insightful questions.
% Developing and cleaning these domain specific KG is not trivial. 
More details of KG construction are provided in the Appendix~\ref{kg-building-and-cleaning}.

\subsection{Metrics}
\label{sec:metrics}
To measure the alignment between KGs and LLMs, here we introduce three edge-level metrics, where each edge is evaluated multiple times with $m$ successes and $n$ failures.

\textbf{Win rate}. For each edge, LLM wins if the number of successes surpasses the number of failures, namely $m > n$.
The win rate signifies the portion of winning edges out of all the examined edges.

\textbf{Zero-sense rate}. An LLM would has zero-sense about an edge~(fact) if the model has never answered the edge correctly, namely $m = 0$.
The zero-sense rate signifies the portion of edges with zero-sense.

\textbf{All-sense rate}. An LLM would has all-sense about an edge~(fact) if the model has never failed to answer the edge, namely $n = 0$.
The all-sense rate signifies the portion of edges with all-sense.

Based on the definition above, win rate is the portion of edges that an LLM has higher chance to answer them correctly, indicating the reliability of LLM.
Zero-sense rate is the portion of edges that an LLM always fails to answer.
All-sense rate is the portion of edges that an LLM always succeed to answer.
% Higher win rate tells us that the LLM is more reliable for the testing KG.
% Higher zero-sense rate tells us the LLM knows less about the testing KG.

\begin{figure}[!t]
\centering
\includegraphics[width=0.3\textwidth]{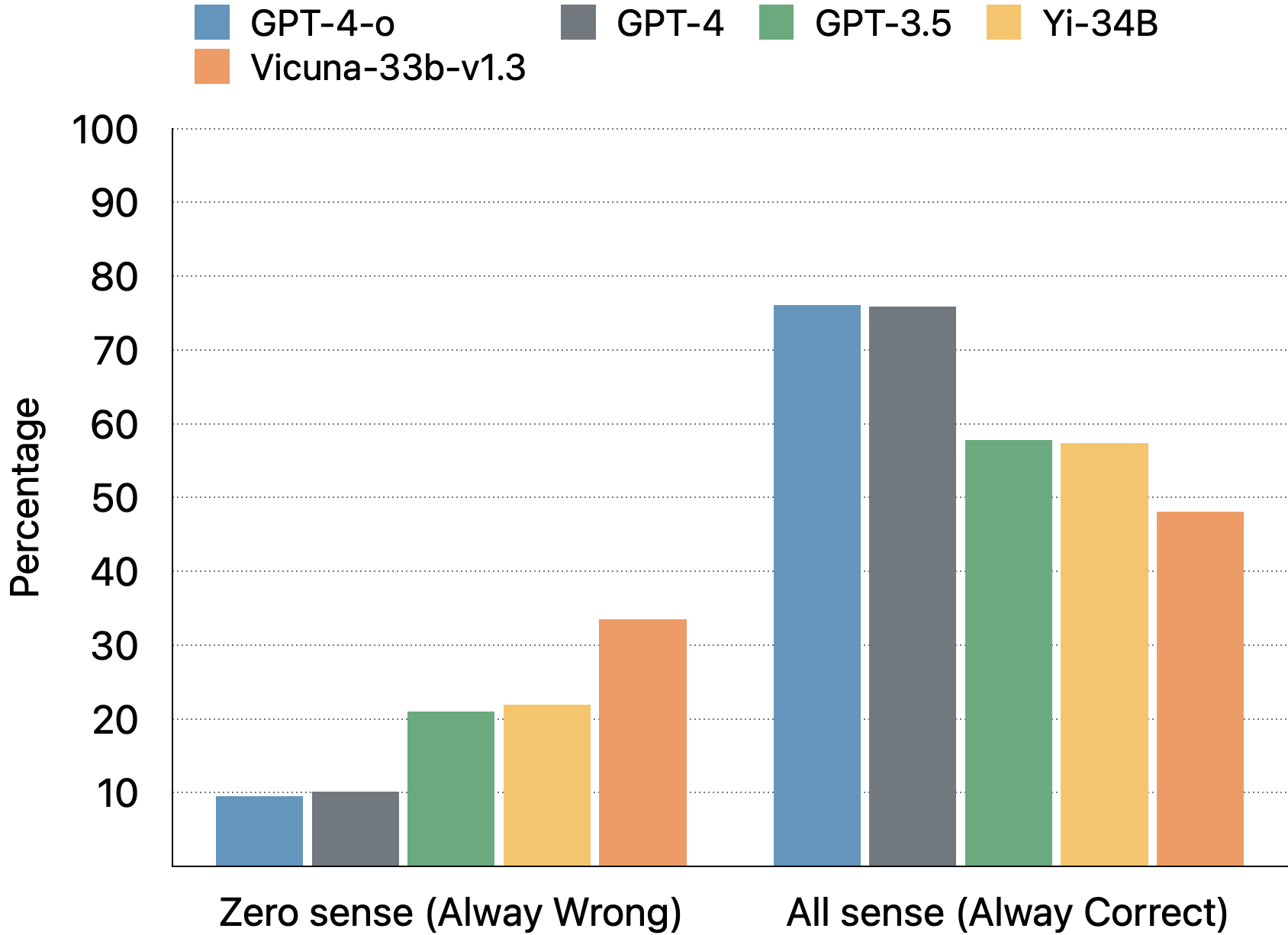}
\caption{Percentage of facts that LLMs always answered correctly and incorrectly. Full results in Tab.~\ref{tab:zero-sense-rate},~\ref{tab:all-sense-rate}.}
\label{fig:zero-all-sense}
\end{figure}

\subsection{Main Results}

To evaluate the KGs, we run \ours~across LLMs with 60 iterations and 64 batch size for each graph.
Tab.~\ref{tab:win-rate} show results of win rate.
We put results of zero-sense rate and all-sense rate in Appendix Tab.~\ref{tab:zero-sense-rate},~\ref{tab:all-sense-rate}, but we visualize 5 models' results in Fig.~\ref{fig:zero-all-sense}.

Across varying difficulty levels, knowledge graphs, and the tested models, GPT-4 family consistently outperforms the others in both metrics. 
The performance across GPT-4 variations are close, where GPT-4, GPT-4o and GPT-4-turbo exhibits the most comparable performance, followed by GPT-4-1106-preview. 
We find GPT-4o exhibits a more cautious disposition in disclosing personal information. 
It frequently refrains from providing specific details about individuals, which may be correlated with its performance disparity in NBA KG, where GPT-4 underperforms GPT-4. 
Further investigation is warranted.
We also find the gap between GPT-3.5-turbo and GPT-4 relatively larger across all domains and all difficulty levels, and GPT-3.5-turbo is even worse than the legacy LLMs under NBA KG EASY mode.
% The exception arises when evaluating against the NBA sub-knowledge graph at the EASY difficulty level, where Babbage-002 and Davinci-002 surpass GPT-3.5-turbo. 
Upon investigating the evaluation logs, GPT-3.5 exhibits a conservative approach, abstaining from generating answers when lacking confidence rather than providing speculative responses. 
% We have reasons to believe that an extra fine-tuning step is implemented to curb undesirable behavior in LLM. 
Responses following this protocol consistently begin with the phrases, ``I am sorry, but I couldn't find any information on/about...'', ``I'm sorry, but as an AI assistant, I do not have the capability to provide real-time information ...''. 
In such cases, the edge would be marked as failed when the model declines to answer a question.
We also observed such behavior in Yi-34B-Chat and Vicunna-33b-v1.3.
% It seems as though GPT-3.5-turbo could have performed better if there were no restrictions in place.

% The open sourced model Yi-34B-Chat is able to achieve similar performance to GPT-3.5-turbo, and it beats GPT-3.5-turbo on NBA KG on both Easy and Hard mode.

% We find the open sourced model Yi-34B-Chat is comparable to the GPT-3.5-turbo model and outperforms GPT-3.5-turbo in the NBA KG dataset on both Easy and Hard modes. This is a remarkable achievement for Yi-34B-Chat, considering its smaller size compared to GPT-3.5-turbo. A smaller-sized LLM called Xwin-LM-13B-V0.2, also did something very interesting. It followed the trend set by Yi-34B-Chat and outperformed Vicuna-33b in our experiments. Given Vicuna-33b-v1.3 is an instruction-fine-tuned model, it only has slightly edged out legacy OpenAI completion models. In fact Vicuna-33b-v1.3 only performs better in answering Country related questions.

Lastly, we find the two legacy models exhibit comparable performance across evaluations.
The random guessing baseline of the win rate is 50\% for EASY evaluation, and 25\% for HARD evaluation. 
We find Babbage-002 and Davinci-002 results are only slightly better than the random guessing, clearly showing the gap between the legacy LLMs and the recent LLMs.
We also provide examples of different error types in Tab.~\ref{tab:case-study}.
% \ours~offers a distinct advantage by providing multiple opportunities for the targeted LLM to be tested against the same edge. 
% Triples that are labeled as zero-sense are highly indicative of blind spots of knowledge within the LLM.

\begin{figure}[!t]
\centering
\includegraphics[width=0.4\textwidth]{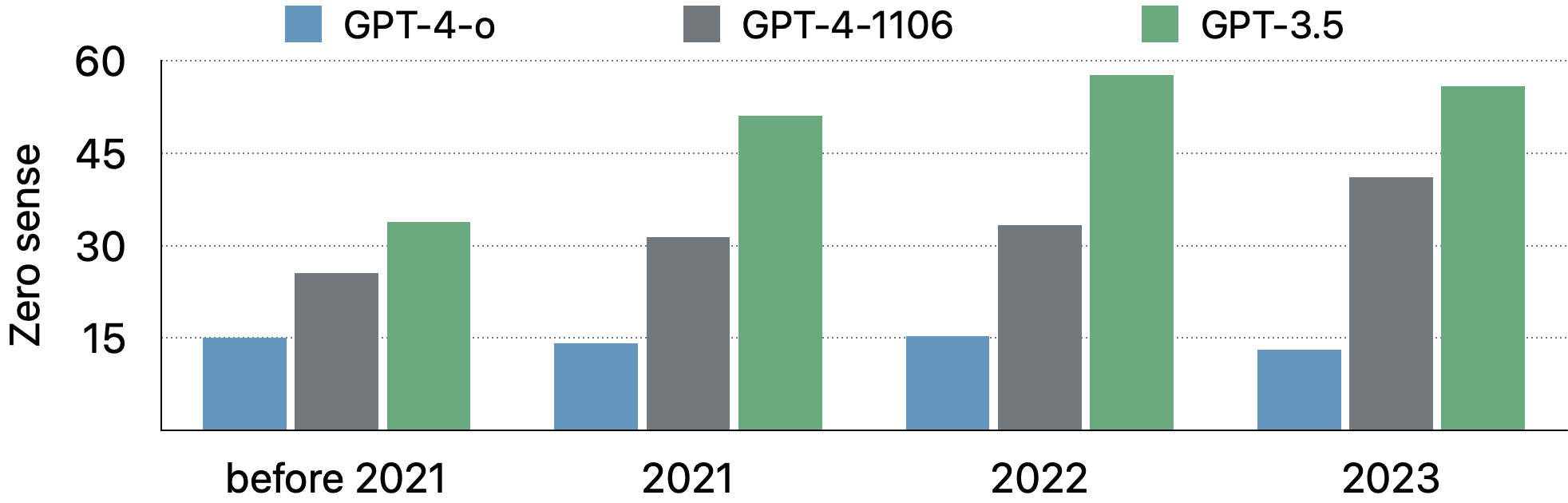}
\caption{Zero-sense rate across years. Full results in Fig.~\ref{fig:movie-hard-years}.}
\label{fig:temporal}
\end{figure}
\subsection{Results Analysis by Edge Attributes}
Another advantage of evaluating LLM with KG is that the results can be aggregated by different edge attributes.
In this section, we show \ours~can be used for two different focuses of evaluation including the temporal groups and entity groups.

\subsubsection{Temporal Analysis of Results} \label{sssec:temporal-group}
We conduct temporal analysis with movie KG.
We group the results by the movie release years in Fig.~\ref{fig:temporal} and full results are in Fig.~\ref{fig:movie-hard-years} in Appendix.
From this Fig.~\ref{fig:temporal}, we observe that both the GPT-3.5 and GPT-4-1106 perform worse for questions after 2020, which is reasonable as they were mainly pretrained with data before September 2021.
On the other hand, we found that GPT-4-o significantly outperforms the other models in terms of zero-sense rate and win rate.
In Fig.~\ref{fig:movie-hard-years}, we find all models get worse when evaluated in HARD mode, but GPT-3.5 is more robust.
This is because a big portion of GPT-3.5's failures are caused by refusing to answer the questions, instead of providing the wrong answers, which explains its results in EASY and HARD testing.
% Interestingly, we find all three recent LLMs perform worse for movies released in 2018, which might related to the pretraining data collection but need further investigation as their pretraining data are not publicly available.
Finally, it should be noted that it is reasonable that the rankings in Fig.~\ref{fig:movie-hard-years} are not strictly aligned with the years, as the temporal difference is not the only factor that affect the evaluation results.

\subsubsection{Entity Groups Analaysis} \label{sssec:entity-group}
We also show results where we group the Country KG edges by the entity type in Fig.~\ref{fig:country-hard} in appendix.

The proficiency levels across countries can be visualized using a color coded table, where a darker color signifies higher zero-sense rate and thus a lower level of proficiency.
Taking GPT-4 evaluated against country KG under HARD level difficulty for example, GPT-4 exhibits a recognition accuracy where the Austria, Mexico, and Italy are identified and ranked as 1, 2, and 3 respectively. In contrast, countries such as Canada, Philippines, and the United Kingdom are positioned at the lower end of the ranking scale.

The rationale behind the ranking can be elucidated by examining the dotted heatmap in the appendix(Fig.~\ref{fig:gpt-4_country_hm}). 
In this figure, the size of each dot corresponds to the number of edges within the predicate sub-group, normalized by the total size of edges in the entire group. 
Additionally, the color of each dot serves as an indicator of the knowledge proficiency associated with the predicate sub-group pertaining to the respective country. Contrary to the table color theme, the darker color here indicates lower zero-sense rate and thus higher level of proficiency.

We find \ours~can identify where the errors came from for each country group.
Concentrating on the Austria and the Canada, which represent the highest and lowest ranked countries, respectively, it becomes evident that GPT-4 exhibits enhanced proficiency pertaining to specific predicate sub-groups. Notably, these sub-groups include ``located in time zone'', ``located in the administrative territorial entity'', ``electrical plug type,'' ``emergency phone number,'' and ``head of state''.

\subsection{Human Evaluation}
\begin{table}[!t]\centering
\scriptsize
\setlength{\tabcolsep}{5pt}
\begin{tabular}{lrrrrr}\toprule
&Country &NBA &Movie &Average \\\midrule
QG &96\% &98\% &99\% &97.7\% \\
- Wh-question &100\% &100\% &98\% &99.3\% \\
- Yes/No-question &92\% &96\% &100\% &96\% \\\midrule
QAV &96\% &96\% &96\% &96\% \\ \midrule
QG+QAV &94\% &96\% &97\% &95.7\% \\
\bottomrule
\end{tabular}
\caption{Human assessment of question generation~(QG) module and question answering verification~(QAV) module. 
% explain the accuracy 
Majority voting among five annotators was employed as the method for rendering a final judgment.}\label{tab:human_eval}
\end{table}
We conduct human evaluation to verify the question generation module and the question answering module of \ours.
A random sample of 300 instances was obtained~(100 per domain, 50 per question type), and human annotations were acquired through five rounds of rating. 
The assessment process is conducted instance by instance, where the annotators were tasked with evaluating two aspects~(QG and QA): firstly, the clarity of the generated question's intent, and secondly, the correctness of the LLM's response in relation to the ground truth answer and its synonymous expressions.
These second objective is to verify if the annotator's judgement agrees with \ours's judgement, and is only conducted for Wh questions, as there is no need to verify Yes/No by human.
After collected the ratings, a majority voting mechanism was employed for each instance, wherein a label was assigned as "True" if at least three annotators concurred on the evaluation criterion.
The evaluation results are presented in Tab.~\ref{tab:human_eval}, and \ours~demonstrates robust performance in human evaluation across domains. 96\% accuracy in QAV and an impressive 98\% accuracy in QG.
We also report the overall accuracy of \ours~.
For the purpose of this evaluation, we define an instance as correct when two conditions are met: the generated question is marked as correct by human; and the QA correctness judged by \ours~aligns with human judgment.
Our findings show a 95.7\% accuracy rate for \ours~, indicating its ability to approximate human-level performance.
\section{Related Work} 
It's an established fact that pre-trained models have the ability to learn and retain knowledge.
For example, \citet{petroni2019language} discovered that BERT~\cite{devlin2018bert}, even without finetuning, harbors relational knowledge comparable to traditional NLP methods. 
With LLMs showcasing superior in-context learning and knowledge retention, evaluating their knowledge becomes pivotal to bolster performance and mitigate hallucination. 

The knowledge assessment often tests the model with specific knowledge-related datasets~\cite{lewis2021paq, petroni2020kilt, roberts-etal-2020-much, peng2023check, press2022measuring, mallen2023not}. 
However, given the fact that LLMs are trained on web-crawled corpora and the data is constantly evolving, it is hard to exclude the test examples from the pretraining corpus.
For example, \citet{deng2023benchmark} use fill-in probing and multi-choice probing to check the data leakage of pretrained LLMs.
Their results show that GPT-3.5-turbo exhibited a noteworthy ability to guess the missing option.
Another concern is that the knowledge is dynamic, and the evaluation datasets remain fixed, which makes it challenging to evaluate the LLMs' knowledge accurately.
\citet{dhingra-etal-2022-time} propose a diagnostic dataset that pairs the text and timestamp together and jointly models text and time.
However, their dataset is static and designed for 2010 to 2020, which is not suitable for evaluating the LLMs' knowledge in the future.
Finally, the predominant metric employed by these datasets revolves around the test set accuracy, making it challenging to identify solutions for enhancing the LLM and reducing the hallucination.

On the other hand, knowledge graphs have the advantages of customization to specific domains, evolving knowledge, and reduced potential for test set leakage, which has been employed as a structured knowledge source for LLMs~\cite{lin2019kagnet, agarwal2020knowledge, rosset2020knowledge} and also been employed as a tool to probe knowledge in LLMs.
LAMA~\cite{petroni2019language} is the first work to probe a pretrained model with KGs, where they use the KG to generate the cloze statement and evaluate the LM's knowledge with accuracy.
However the cloze statement is not a natural question, and the correct answer is not unique in many cases, making the evaluation inaccurate.
LPAQA~\cite{jiang2020can} propose to mine the relation words from the web for each subject-object pair, which is impractical for large knowledge graph.
In addition, these methods mainly focus on the accuracy but neglect that LLMs may respond differently to the same fact, where reliability should also be considered.
KaRR~\cite{dong2023statistical} proposes to solve this issue by using multiple prompts for each KG edge and using the output logits of LLMs to measure the knowledge reliability.
However, KaRR is inefficient for large graphs, and it is not generalizable due to the unavailable of LLM's output logits.
Moreover, transforming KG triplets into questions is more natural than the text cloze task, but previous works mainly adopt the text cloze task for simplicity. 
%the structure of the KG, for example, the neighbors, alias, out/in degree, etc., which can be helpful in generating the questions and verifying the answers more accurately.
Finally, to our best knowledge, there is no existing work that visualizes the LLM's knowledge with KG~(Fig.~\ref{fig:pkg}).
\section{Conclusion}

In this work, we introduced \ours, a novel and efficient method tailored for visualizing and evaluating the factual knowledge embedded in LLMs. 
Our proposed Thompson sampling inspired framework with parameterized KG offers a more efficient way to revil LLM's knowledge blind spots than existing brute force iterating methods. 
By evaluating various LLMs with our developed domain-specific KGs, we show \ours~provides adaptable and customizable views of an LLM's knowledge.
% In addition to evaluating the accuracy of facts, our proposed parameterized KG offers an efficient way to assess the knowledge reliability of LLMs. 
Human evaluation results indicate that \ours~can access LLMs with a level of accuracy nearly qeuivalent to that of human annotators, achieving 95.7\% of the accuracy rate.
Furthermore, our tool~\ours, together with our assessment KGs, sourced from Wikidata, will be available to the research community, fostering collaboration and serving as a valuable resource for future investigations into language models.
For businesses employing LLMs, our contributions facilitate the development of more reliable and factual AI systems, fostering trustworthy user experiences and efficient processes for improving model knowledge.

\section{Limitation}

KG plays a pivotal role in our approach, and its quality significantly impacts the effectiveness of this method. A high-quality KG is essential not only for the Question Generation step to generate meaningful questions but also for signal propagation. If the KG is fragmented and scattered, signal propagation then becomes less beneficial.

While our current method incorporates counting updates for alpha and beta, we acknowledge the potential for improvement. Exploring alternative methods for updating these parameters is an area of active research for us. 

The signal propagation method is another direction that we can dive into, instead of only propagate to neighbour edges, should we also propagate to further edges? Instead of equally update the neighbour edges, should we decay the signal? etc.

Question generation currently is limited to just one hop, being able to generate complicated questions that evolves multiple edge hops would enable our method to evaluation the model not only on factual knowledge retrieval, but also complex reasoning capability. 

\section{Ethical Considerations}
We foresee no ethical issues originating from this work.

% \balance
\bibliography{custom}
\bibliographystyle{acl_natbib}

\clearpage
\newpage
\appendix
\onecolumn
\begin{center}
    {\Large \bf Appendix}
\end{center}

\section{Appendix}
\label{sec:appendix}
\subsection{Cost Analysis}
Here we highlight that the cost of GPT-4 is not counted by the number of queries, but the number of tokens. After doing a cost analysis, we conclude \textit{GPT-4 only cost about \$20 per graph}, which is acceptable. 

GPT-4 in \ours~is only used to construct the question and verify LLM answers, and all the generation is based on knowledge graph triplets. And for yes/no questions, we simply use string matching to verify the answer.
Our QG prompt using GPT-4 is around 60-100 input tokens + 10-30 output tokens, and 60 input token plus 8 output tokens for answer verification. 
Currently based on OpenAI website, gpt-4-0125-preview and gpt-4-1106-preview costs \$10 per 1M input tokens and \$30 per 1M output tokens. If we take an upper bound cost of our request 100 input tokens and 30 output tokens, each request costs <\$0.002. The final cost of course also depends on the size of the knowledge graph, but in our case for example, NBA knowledge graph has 2689 active edges, 22.5K API requests is sufficient to iterate over the entire edge set 8 times, and this translates to \$45 (in reality if using \ours, using less than half the number of API requests can get to a decent theta estimation \$20 per evaluation)

The major cost is actually the hosting cost of the target LLM. For reference 1 A100 GPU cost is around \$5 per hour, and this cost scale up easily when evaluating larger LLMs.

In addition, we believe for answer verification step, it is not a requirement to use GPT-4 (we choose it for simplicity), maybe a lower cost model could be used to replace GPT-4.

\subsection{Knowledge Graph Building and Cleaning}
\label{kg-building-and-cleaning}
% The graph building begins with a set of human-defined central points~(such as country names, NBA team names), from which random walks are initiated to growing the KG.
% % This strategic sampling allows us to maintain the edge distribution.
% After that, entity and predicate filters are applied. 
% We keep entities mentioned in multiple languages, as they often hold broader significance. 
% Frequently occurring entities are also given prominence due to their centrality within various contexts. 
% Additionally, we exclude entities without aliases to focus on well-recognized and frequently referenced entities. 
% The number of predicates in a KG is not significant, so we manually examine the KG predicates to exclude trivial, malformed, or ambiguous ones.

% As a result the parameterized KGs concentrates on specific domains and contains only relevant, significant knowledge as defined above. 

Given Wikidata's vastness and inherent noise, we implement multiple strategies to maintain focus, relevance, and precision. Following techniques empower us to delve into specialized domains and ensure us a targeted and reliable exploration of the data.

\subsubsection{Sampling Strategies and Preserving Data Distribution}
Maintaining the original data distribution is important when cleaning a knowledge graph. To achieve this, random walk with both forward \ref{subsec:forward-walk-query} and backward \ref{subsec:backward-walk-query} dimension are employed. Sorting by random value of each queried edges, the sub-knowledge graph contains the representative samples that mirror the diversity of the original knowledge graph, we can preserve the inherent distribution of entities and relationships. This approach guarantees that our cleaned knowledge graph remains a faithful representation of the underlying data, enabling us to draw accurate conclusions from our research.

The extent of the random walk distance is flexible and tailored to specific requirements. Within our sub knowledge graphs, we conduct random walks spanning three steps, encompassing both nodes and edges within this range for analysis.

\subsubsection{Focus and Curated Relevance}
In the realm of knowledge graphs, Wikidata stands out as a repository of extensive information. However, our research necessitates a more nuanced approach. While Wikidata offers comprehensive knowledge, our focus lies in curated topics and entities tailored for specific purposes. This distinction is vital as it allows us to delve deeper into specialized domains, ensuring the precision and relevance of the data we analyze.

To address this issue, the parameterized knowledge graph begins by establishing a set of human selected central entities, from which it initiates random walks to explore neighboring entities. Additionally, we perform predicate analysis to discern and exclude predicates of lesser importance or those that are overly common. This approach ensures the focus on pertinent data while filtering out less relevant information.

\subsubsection{Filtering Less Relevant Entities}
The other challenges we encounter in Wikidata pertains to the noise within its knowledge graph. This noise manifests in the form of entities that are rarely mentioned or of lesser importance in the context of our research objectives. To maintain the integrity of our analysis, it is important to identify and filter out these less relevant entities.
\begin{itemize}
    \item Filtering by language count: entities mentioned in multiple languages are often more significant and relevant to a broader audience. By focusing on such multilingual entities, we ensure the inclusion of globally relevant information in our analysis.
    \item Filtering by word frequency: entities that are frequently mentioned in various contexts are likely to hold greater importance. By considering word frequency, we prioritize entities that are central to discussions, thereby enhancing the relevance and significance of the data included in our analysis.
    \item Filtering out entities with no alias: entities without aliases are less likely to be widely recognized or referenced. By excluding these entities, we focus our analysis on well-known and frequently mentioned entities, aligning our research with more meaningful and impactful data points.
\end{itemize}

\subsection{Uncovered Error Types}
\begin{table*}[h]
\small
\centering
\setlength{\tabcolsep}{4pt}
\begin{tabular}{m{0.2\textwidth}m{0.25\textwidth}m{0.25\textwidth}m{0.15\textwidth}}
\toprule
\textbf{Error Type} & \textbf{Question} & \textbf{Response} & \textbf{Answer} \\
\cmidrule{1-4}
\textbf{Factual Error} & On which side does the railway traffic run in Israel? & Railway traffic in Israel runs on the right-hand side. & Left side \\
\cmidrule{1-4}
\textbf{Obsolete Knowledge} & Who is the current head of state in England? & The current head of state in England is Queen Elizabeth II. & Charles III of the United Kingdom \\
\cmidrule{1-4}
\textbf{Self-contradiction} & Is the Australian dollar the currency of Nauru (also known as NR)? & No, the Australian dollar (AUD) is the official currency of Nauru, a small island nation in the Pacific Ocean. & Yes \\
\cmidrule{1-4}
\textbf{Inconsistent Response} & In Tonga (also known as TO), do people drive on the right side of the road? & No, in Tonga, people drive on the left side of the road. & No \\
& Is the left the driving side in Tonga (also known as TO)? & No, the right is the driving side in Tonga. & Yes \\
\bottomrule
\end{tabular}
\caption{Error types uncovered from the country KG.}
\label{tab:case-study}
\end{table*}

% \subsection{GPT-4o Blind Spots}
% \begin{table*}[h]
% \small
% \centering
% \setlength{\tabcolsep}{4pt}
% \begin{tabular}{m{0.2\textwidth}m{0.25\textwidth}m{0.25\textwidth}m{0.15\textwidth}}
% \toprule
% \textbf{Fact} & \textbf{Question} & \textbf{Response} & \textbf{Answer} \\
% \cmidrule{1-4}
% (Bob Woollard, drafted by, New York Knicks) & Was Bob Woollard (also known as Robert George Woollard) drafted by the New York Knicks? & No. & Yes. \\
% \cmidrule{1-4}
% Obsolete Knowledge & Who is the current head of state in England? & The current head of state in England is Queen Elizabeth II. & Charles III of the United Kingdom \\
% \cmidrule{1-4}
% Self-contradiction & Is the Australian dollar the currency of Nauru (also known as NR)? & No, the Australian dollar (AUD) is the official currency of Nauru, a small island nation in the Pacific Ocean. & Yes \\
% \cmidrule{1-4}
% Inconsistent Response & In Tonga (also known as TO), do people drive on the right side of the road? & No, in Tonga, people drive on the left side of the road. & No \\
% & Is the left the driving side in Tonga (also known as TO)? & No, the right is the driving side in Tonga. & Yes \\
% \bottomrule
% \end{tabular}
% \caption{Error types uncovered from the country KG.}
% \label{tab:case-study}
% \end{table*}

\subsection{Zero-sense rate and all-sense rate}
\begin{table*}[!htp]\centering \scriptsize
   \begin{tabular}{lrrrrrrrr}\toprule
    \multirow{2}{*}{LLMs} &\multicolumn{2}{c}{Country} &\multicolumn{2}{c}{NBA} &\multicolumn{2}{c}{Movie} &\multirow{2}{*}{Average} \\\cmidrule{2-7}
    &EASY &HARD &EASY &HARD &EASY &HARD & \\\midrule
    Babbage-002 &24.51 &51.56 &15.34 &38.61 &26.70 &56.77 &35.58 \\
    Davinci-002 &24.44 &47.27 &17.69 &37.89 &28.54 &52.71 &34.76 \\
    Vicuna-33b-v1.3 &17.19 &26.09 &41.75 &37.30 &36.01 &42.47 &33.47 \\
    Xwin-LM-13B-V0.2 &28.96 &35.12 &19.06 &26.92 &34.59 &38.48 &30.52 \\
    Yi-34B-Chat &16.15 &25.17 &14.16 &18.79 &26.58 &30.90 &21.96 \\
    GPT-3.5-turbo &14.98 &20.32 &17.17 &21.09 &22.70 &29.36 &20.94 \\
    GPT-4-1106-preview &7.59 &14.16 &8.19 &12.42 &9.21 &21.43 &12.17 \\
    GPT-4-turbo &7.26 &12.73 &6.77 &9.44 &8.51 &17.52 &10.37 \\
    GPT-4 &7.42 &12.99 &6.07 &\textbf{8.13} &8.35 &17.67 &10.11 \\
    GPT-4o &\textbf{5.84} &\textbf{11.86} &\textbf{7.43} &8.24 &\textbf{7.63} &\textbf{16.33} &\textbf{9.56} \\
   \bottomrule
   \end{tabular}
   \caption{Zero-sense rate results for different LLMs evaluated under EASY and HARD modes.}\label{tab:zero-sense-rate}
\end{table*}

\begin{table*}[!htp]\centering \scriptsize
   \begin{tabular}{lrrrrrrrr}\toprule
    \multirow{2}{*}{LLMs} &\multicolumn{2}{c}{Country} &\multicolumn{2}{c}{NBA} &\multicolumn{2}{c}{Movie} &\multirow{2}{*}{Average} \\\cmidrule{2-7}
    &EASY &HARD &EASY &HARD &EASY &HARD & \\\midrule
    Babbage-002 &54.09 &33.63 &43.54 &23.98 &53.96 &29.91 &39.85 \\
    Davinci-002 &55.70 &37.16 &43.87 &25.93 &52.73 &33.93  &41.55 \\
    Vicuna-33b-v1.3 &64.27 &54.95 &35.05 &39.41 &48.99 &45.64 &48.05 \\
    Xwin-LM-13B-V0.2 &53.93 &48.81 &46.60 &45.68 &52.26 &47.10 &49.06 \\
    Yi-34B-Chat &64.22 &55.45 &56.55 &55.60 &57.97 &54.34 &57.36 \\
    GPT-3.5-turbo &68.43 &60.65 &50.10 &51.33 &59.89 &56.07 &57.75 \\
    GPT-4-1106-preview &79.32 &69.64 &71.78 &65.29 &80.39 &64.47 &71.81 \\
    GPT-4 &82.24 &71.19 &77.44 &73.63 &82.44 &68.58 &75.92 \\
    GPT-4-turbo &80.60 &71.21 &\textbf{75.93} &72.84 &83.42 &71.91 &75.99 \\
    GPT-4o &\textbf{82.95} &\textbf{71.86} &71.98 &\textbf{73.34} &\textbf{83.91} &\textbf{72.81} &\textbf{76.14} \\
   \bottomrule
   \end{tabular}
   \caption{All-sense rate results for different LLMs evaluated under EASY and HARD modes.}\label{tab:all-sense-rate}
\end{table*}

\subsection{Human Evaluation} \label{subsec:human-eval-appendix}
We conduct human evaluation with an internal paid crowdsouring service, where 5 annotators participated the annotation process with their consent of using the data.
All the annotators are from English speaking countries.
The annotation instruction is shown below.

\begin{center}
\begin{tcolorbox}[title=Instruction, width=0.99\textwidth] \label{human-eval-instruction}
In this survey, you are provided with a question, the answer (and answer aliases separated by comma), and an AI's response to this question. 

Background

This survey is to check if the question is understandable, and if the AI's response is matched with the given answer.

Annotation Guideline

As an annotator, you are asked to judge whether the intent of the question is clear enough to be understood (if you can understand the question, please select the option `Yes', whether you know the answer or not).

For some questions, you won't see the answer and AI's response.

For some questions, you will see the answer and AI's response. Then you need to check if the AI's response is correct or not. Please select the option "correct" when the answer, or any of the answer alias appears in AI's response. Otherwise, select ``Incorrect".

Note: You don't need to do any research to find the answer. Just use the provided answer. 
\end{tcolorbox}
\end{center}

\subsection{Prompt} \label {subsec:prompt}
\begin{center}
\begin{tcolorbox}[title=Answer Verification, width=0.99\textwidth] \label{judge-signal}
\textit{Given a question and its correct answers, please judge whether a response is correct or not, only reply with yes or no. \\
\textcolor{blue}{User}: \\
QUESTION: What writing system does the Ixcatec language use? \\
ANSWERS: latin script, latn, roman script \\
\textcolor{blue}{Assistant}: \\
RESPONSE: the ixcatec language uses the latin alphabet for writing. \\
\textcolor{blue}{User}: \\
QUESTION: On which side does railway traffic run in the philippines? \\
ANSWERS: left, left side, left-hand side \\
\textcolor{blue}{Assistant}:
RESPONSE: Railway traffic in the philippines runs on the right-hand side. \\
\textcolor{blue}{User}: \\
QUESTION: <question> \\
ANSWERS: <answers> \\
}
\end{tcolorbox}
\end{center}

\begin{center}
\begin{tcolorbox}[title=Yes/No Question Generation, width=0.99\textwidth] \label{yesno-question-generation}
\textit{
    Given a semantic triple (SUBJECT, PREDICATE, OBJECT), generate a QUESTION that asks if the SUBJECT's PREDICATE is the OBJECT, or includes the OBJECT. The alias of the SUBJECT can be used in the question. \\
    \textcolor{blue}{User}: SUBJECT is Joroud, PREDICT is place of birth, OBJECT is Gamla Uppsala, SUBJECT\_ALIAS is J\"orundr \\
    \textcolor{blue}{Assistant}: Is Gamla Uppsala the birth place of Jorund (also known as J\"orundr)? \\
    \textcolor{blue}{User}: SUBJECT is Asia, PREDICT is located in time zone, OBJECT is Nepal Standard Time, SUBJECT\_ALIAS is Asian continent \\
    \textcolor{blue}{Assistant}:  Is Nepal Standard Time included in the time zones of Asia (also known as the Asian continent)?
    \textcolor{blue}{User}: <s, p, o, [subject alias]>
}
\end{tcolorbox}
\end{center}

\begin{center}
\begin{tcolorbox}[title=Wh-Question Generation, width=0.99\textwidth] \label{generative-question-generation}
\textit{
    Given a semantic triple SUBJECT, PREDICATE, OBJECT, generate a QUESTION that can be answered by the OBJECT. The OBJECT must not occur in the question. \\
    \textcolor{blue}{User}: SUBJECT is Jorund, PREDICT is place of birth, OBJECT is Gamla Uppsala \\
    \textcolor{blue}{Assistant}: What is the birth place of the legendary Swedish king Jorund?\\
    \textcolor{blue}{User}: SUBJECT is Yellow Emperor, PREDICT is father, OBJECT is Shaodian, SUBJECT\_ALIAS \\
    \textcolor{blue}{Assistant}:  Who is the father of Yellow Emperor? \\
    \textcolor{blue}{User}: <s, p, o>
}
\end{tcolorbox}
\end{center}

\begin{center}
\begin{tcolorbox}[title=Yes No Question Answer, width=0.99\textwidth] 
You are a helpful assistant, please answer Yes or No to the user's questions. \\
    \textcolor{blue}{User}:Is Belgium located in the continent of Europe?\\
    \textcolor{blue}{Assistant}: Yes.\\
    \textcolor{blue}{User}: Is Andrzej Duda the head of state of Belgium? \\
    \textcolor{blue}{Assistant}:  No. \\
    \textcolor{blue}{User}: <yes no question>
\end{tcolorbox}
\end{center}

\begin{center}
\begin{tcolorbox}[title=Generative Question Answer, width=0.99\textwidth]
You are a helpful assistant, please give short and accurate answers to the user's question. If there are multiple answers, please list as much as possible. \\
    \textcolor{blue}{User}:What is the birth place of Jorund?\\
    \textcolor{blue}{Assistant}: Gamla Uppsala.\\
    \textcolor{blue}{User}: Who is the father of Yellow Emperor? \\
    \textcolor{blue}{Assistant}:  Shaodian \\
    \textcolor{blue}{User}: <generative question>
\end{tcolorbox}
\end{center}

\subsection{Wikidata Web Query} 
\subsubsection{Forward Walk} \label{subsec:forward-walk-query}
\begin{lstlisting}[breaklines]
SELECT DISTINCT ?subject ?subjectLabel ?subjectDesc ?predicate ?predicateLabel ?predicateDesc ?object ?objectLabel ?objectDesc
WHERE {{
  VALUES ?subject {{ 
    {values}
  }}
  ?subject ?predicate ?object .
  ?subject rdfs:label ?subjectLabel .
  ?subject schema:description ?subjectDesc .
  ?property wikibase:directClaim ?predicate .
  ?property rdfs:label ?predicateLabel .
  ?property schema:description ?predicateDesc .
  ?object rdfs:label ?objectLabel .
  ?object schema:description ?objectDesc .
  FILTER (lang(?subjectLabel) = "en")
  FILTER (lang(?subjectDesc) = "en")
  FILTER (lang(?predicateLabel) = "en")
  FILTER (lang(?predicateDesc) = "en")
  FILTER (lang(?objectLabel) = "en")
  FILTER (lang(?objectDesc) = "en")
}}
ORDER BY UUID()
LIMIT {limit}
\end{lstlisting}

\subsubsection{Backward Walk} \label{subsec:backward-walk-query}
\begin{lstlisting}[breaklines]
SELECT DISTINCT ?subject ?subjectLabel ?subjectDesc ?predicate ?predicateLabel ?predicateDesc ?object ?objectLabel ?objectDesc
WHERE {{
  VALUES ?object {{ 
    {values}
  }}
  ?subject ?predicate ?object .
  ?subject rdfs:label ?subjectLabel .
  ?subject schema:description ?subjectDesc .
  ?property wikibase:directClaim ?predicate .
  ?property rdfs:label ?predicateLabel .
  ?property schema:description ?predicateDesc .
  ?object rdfs:label ?objectLabel .
  ?object schema:description ?objectDesc .
  FILTER (lang(?subjectLabel) = "en")
  FILTER (lang(?subjectDesc) = "en")
  FILTER (lang(?predicateLabel) = "en")
  FILTER (lang(?predicateDesc) = "en")
  FILTER (lang(?objectLabel) = "en")
  FILTER (lang(?objectDesc) = "en")
}}
ORDER BY UUID()
LIMIT {limit}
\end{lstlisting}
\clearpage
\subsection{Additional Figures}

\begin{figure*}[!ht]
   \centering
   \includegraphics[width=1\linewidth]{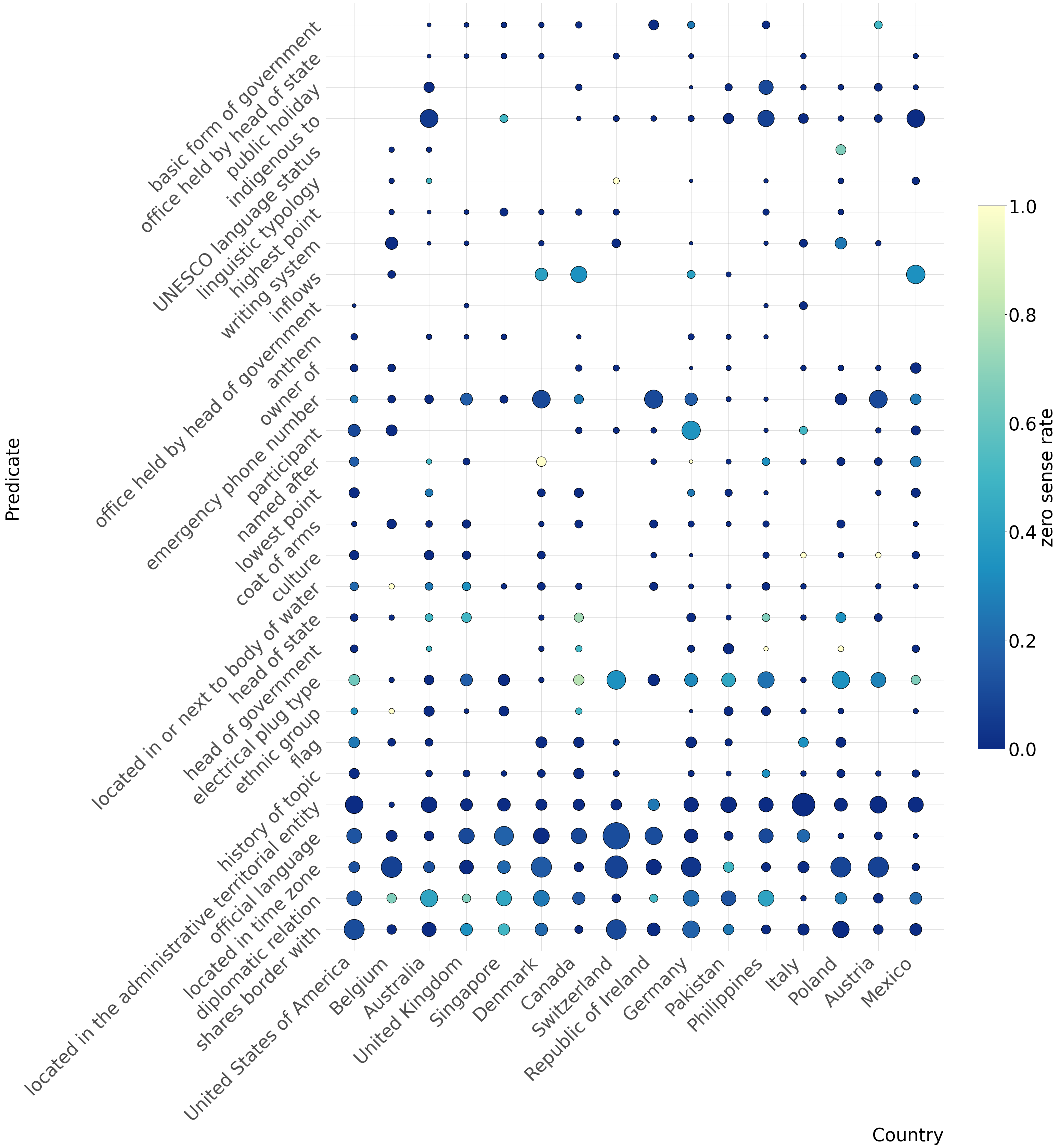}
   \caption{Predicate level knowledge proficiency of GPT-4 evaluated under HARD difficulty.
   The darker color indicates a lower zero-sense rate.
   The dot size shows the proportional size of the number of edges in the predicate sub-group.}
   \label{fig:gpt-4_country_hm}
\end{figure*}

\begin{figure*}[!ht]
  \centering
  \includegraphics[width=1\linewidth]{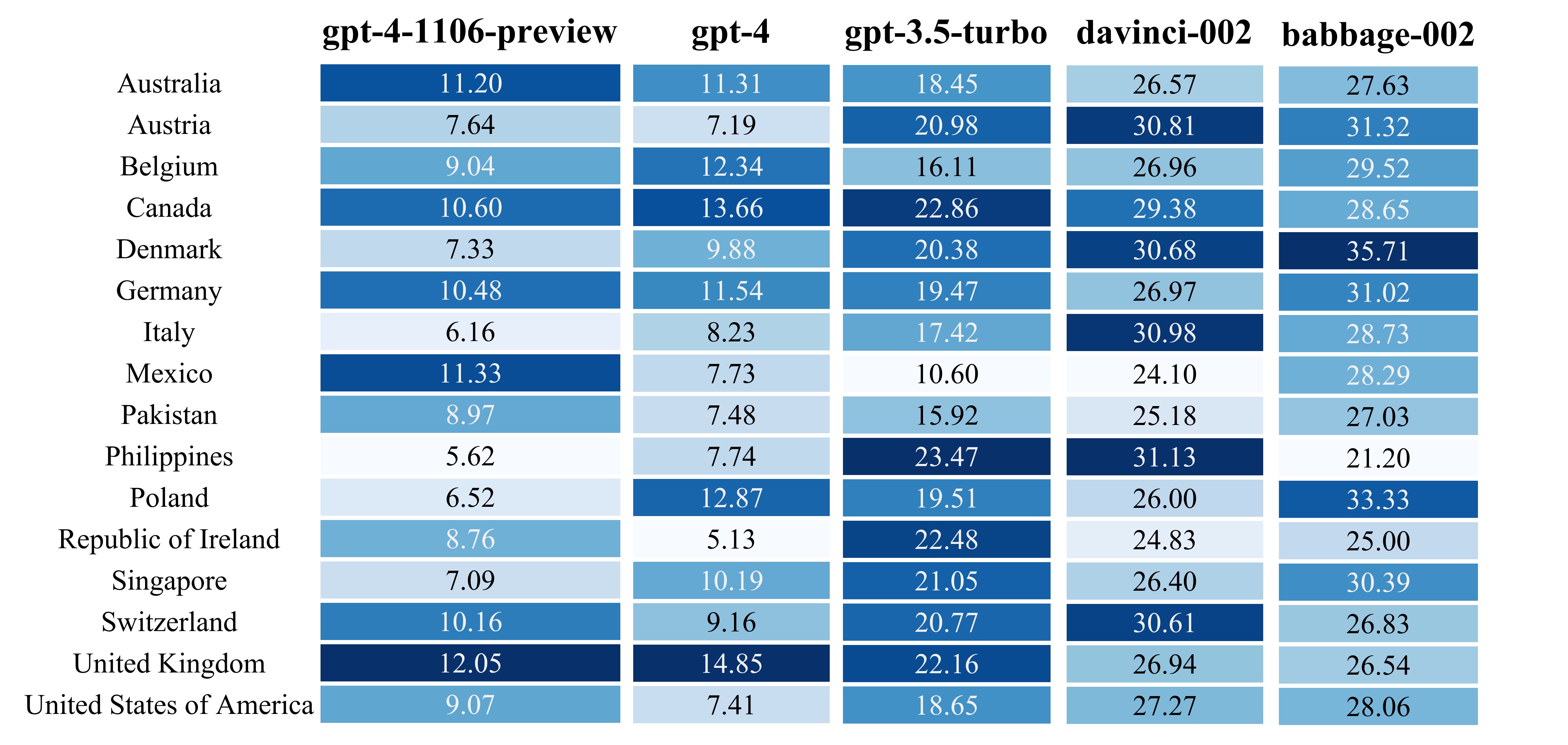}
  \caption{Country KG EASY-level zero-sense rate grouped by countries.}
  \label{fig:country-easy}
\end{figure*}

\begin{figure*}[!ht]
   \centering
   \includegraphics[width=1\linewidth]{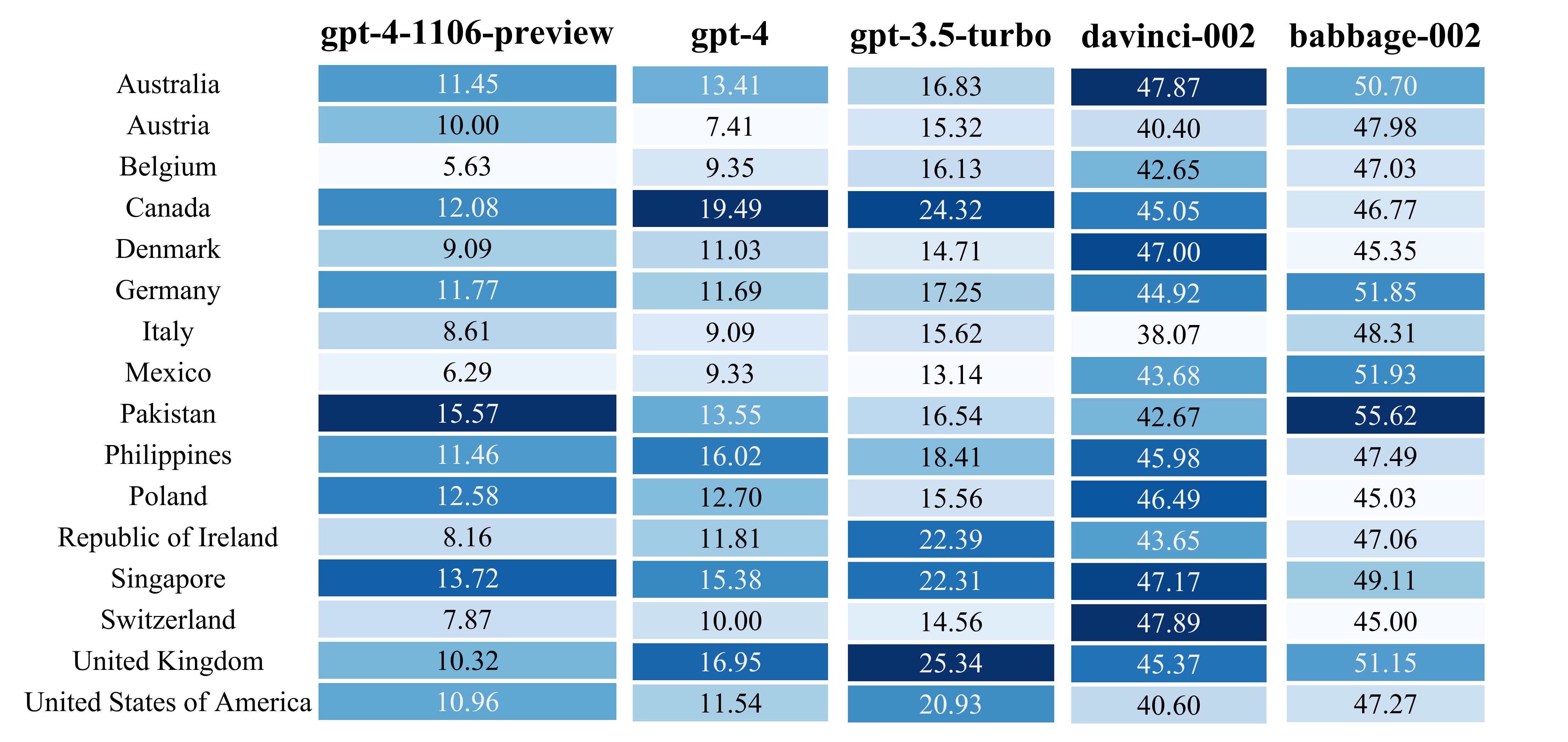}
   \caption{Country KG HARD-level zero-sense rate grouped by countries.}
   \label{fig:country-hard}
\end{figure*}

\begin{figure*}[!ht]
  \centering
  \includegraphics[width=1\linewidth]{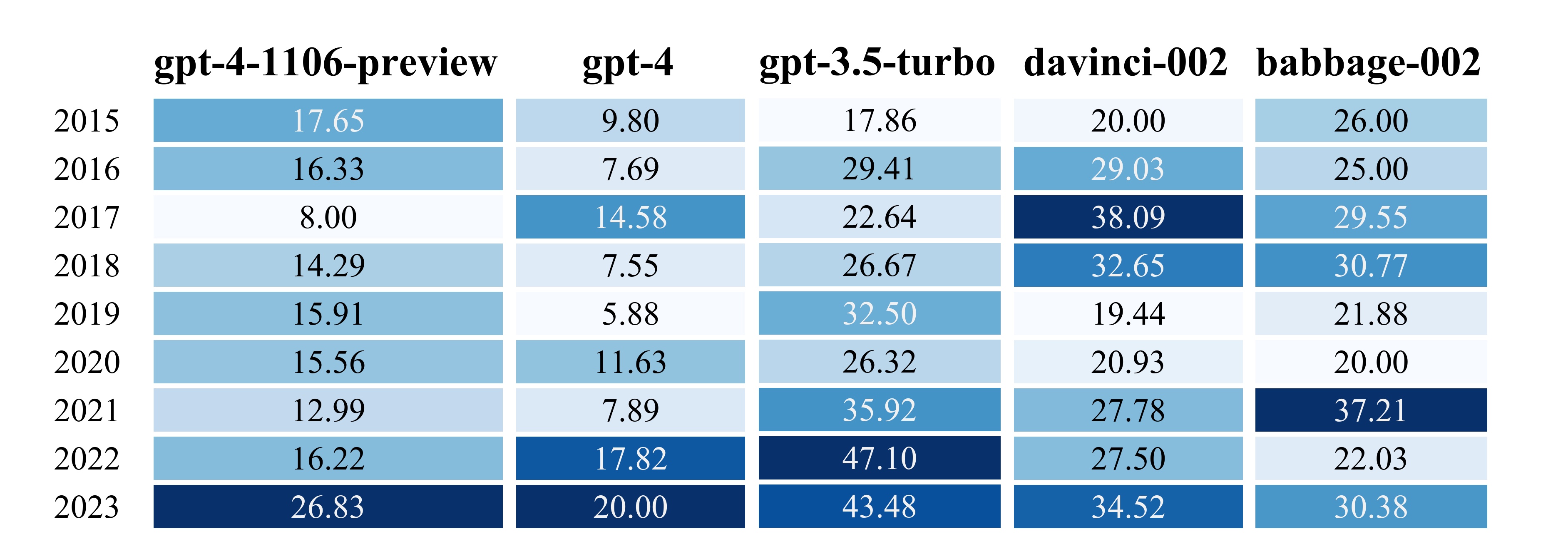}
  \caption{Movie KG EASY-level zero-sense rate grouped by years.}
  \label{fig:movie-easy-years}
\end{figure*}

\begin{figure*}[!ht]
   \centering
   \includegraphics[width=0.8\linewidth]{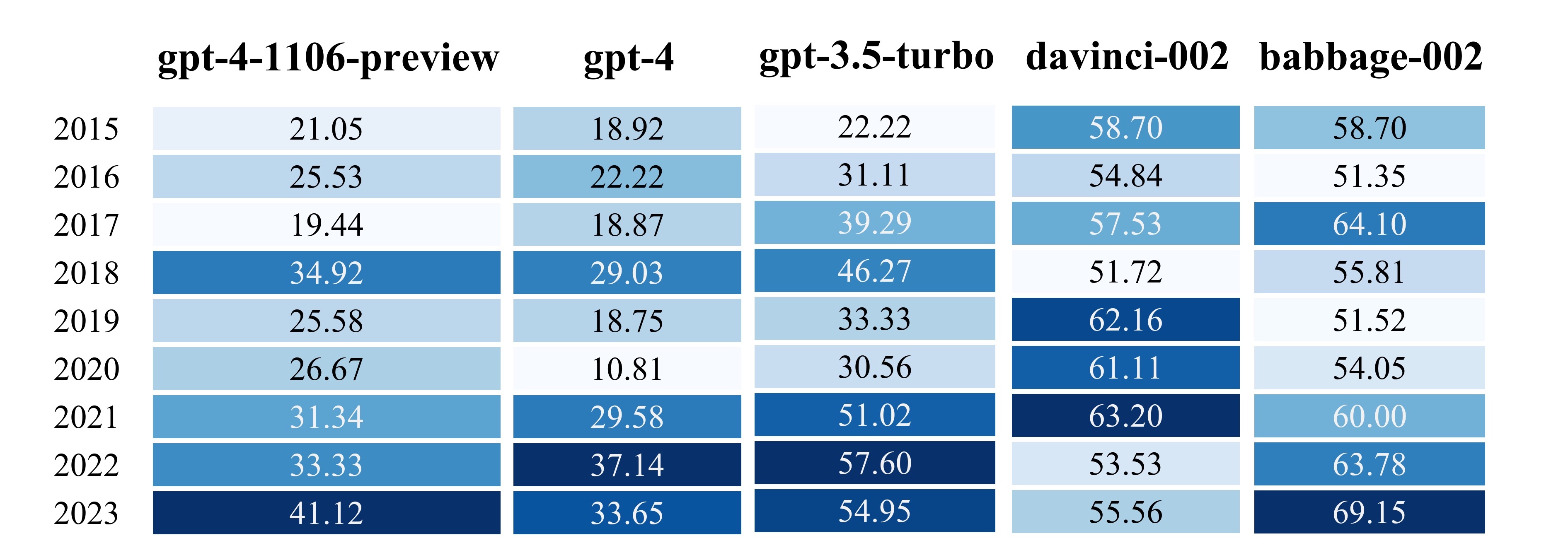}
   \caption{Zero-sense rate grouped by years for moive KG in HARD mode. } 
   % The three recent LLMs perform worse for knowledge after 2020, while the behaviors of the two legacy LLMs are more randomly.
   \label{fig:movie-hard-years}
\end{figure*}

\begin{figure}[!ht]
  \centering
  \includegraphics[width=1\linewidth]{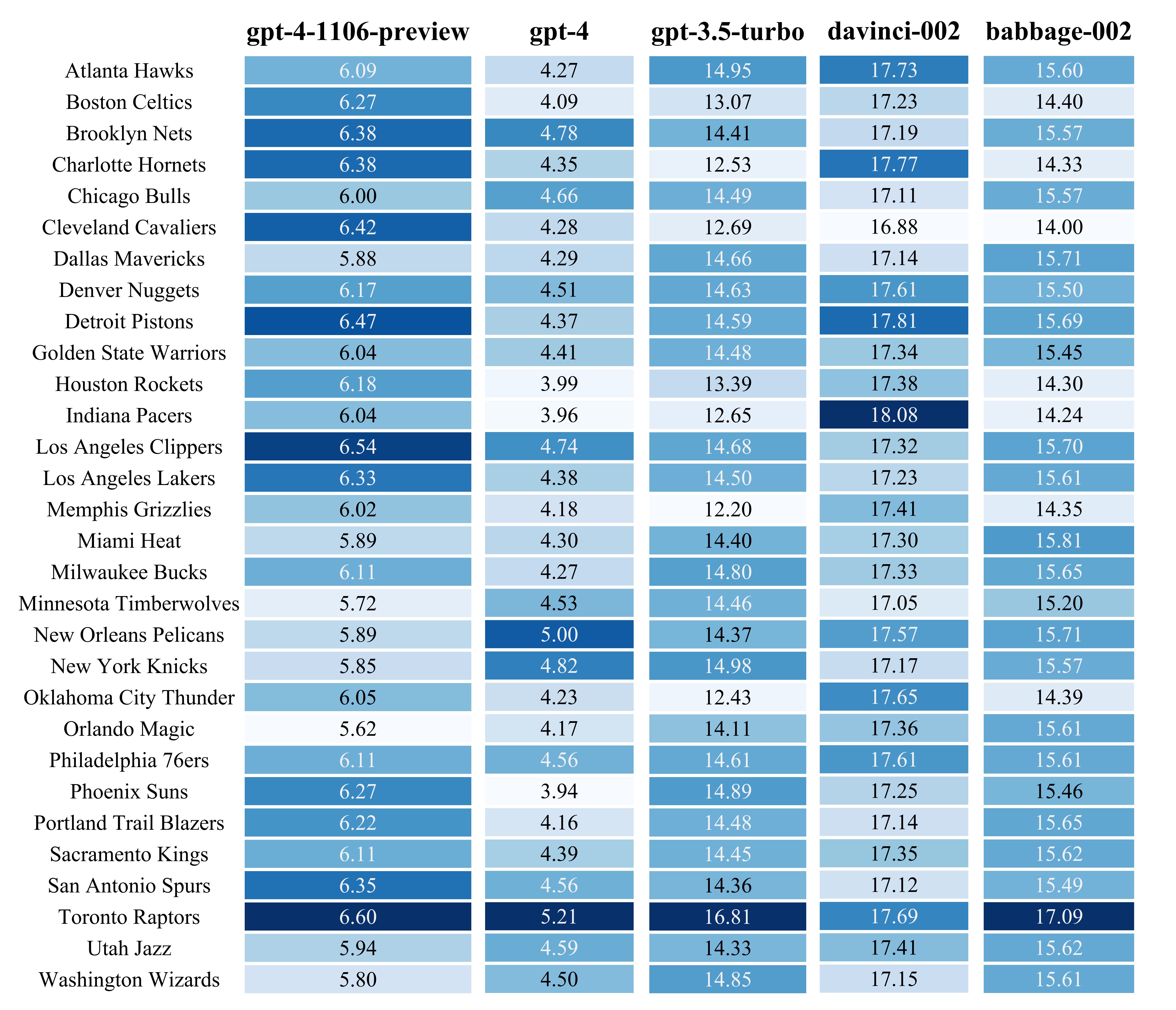}
  \caption{NBA EASY-level zero-sense rate grouped by teams}
\end{figure}
\begin{figure}[!ht]
  \centering
  \includegraphics[width=1\linewidth]{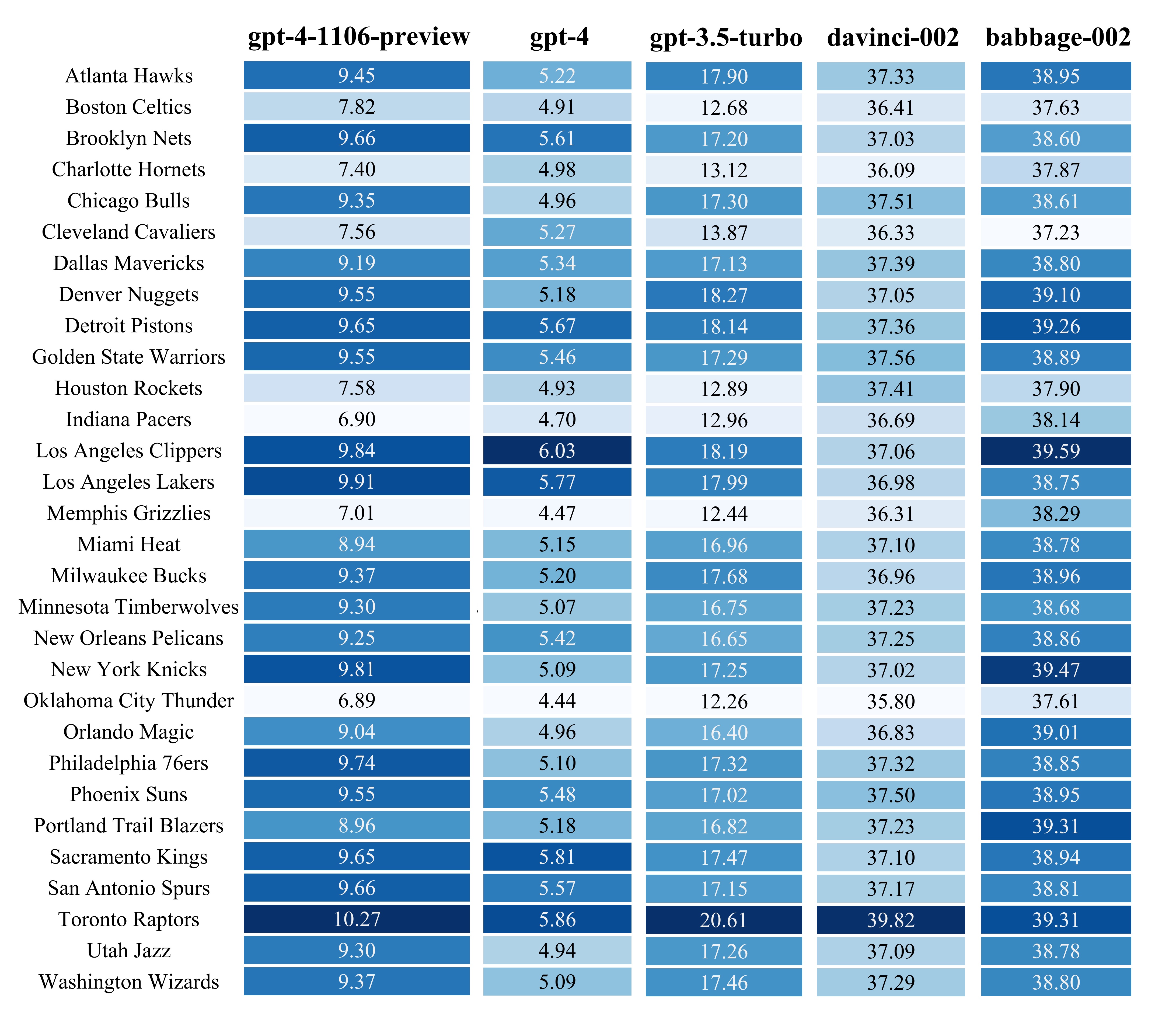}
  \caption{NBA HARD-level zero-sense rate grouped by teams}
\end{figure}

\end{document}